\documentclass[runningheads]{llncs}

\usepackage[T1]{fontenc}

\usepackage{graphicx,verbatim}
\usepackage{amsmath}
\usepackage{amssymb}
\usepackage{multirow}
\usepackage{multicol}
\usepackage{booktabs}
\usepackage[table]{xcolor}
\usepackage[hidelinks]{hyperref} 
\usepackage[nameinlink,capitalise,noabbrev]{cleveref}
\usepackage{pifont}
\usepackage{threeparttable}

\usepackage{marvosym}

\crefname{figure}{Fig.}{Figs.}
\crefname{table}{Tab.}{Tabs.}
\crefname{section}{Sec.}{Secs.}
\crefname{equation}{Eq.}{Eqs.}

\Crefname{figure}{Figure}{Figures}
\Crefname{table}{Table}{Tables}
\Crefname{section}{Section}{Sections}
\Crefname{equation}{Equation}{Equations}

\newsavebox\CBox
\newcommand*\textBF[1]{\sbox\CBox{#1}\resizebox{\wd\CBox}{\ht\CBox}{\textbf{#1}}}

\newcommand{\sig}[2]{%
  \mbox{#1}%
  \rlap{$^{{\fontsize{8pt}{9.6pt}\selectfont #2}}$}%
}

\newcommand{\corremail}{\textsuperscript{\Letter}}

\begin{document}

\title{CSWinUNETR: Segmentation of Thin Anatomical Structures in Medical Images}
\titlerunning{Segmentation of Thin Anatomical Structures in Medical Images}

\author{Junho Moon\inst{1}\orcidID{0009-0004-3522-6357} \and
Haejun Chung\inst{1}\corremail\orcidID{0000-0001-8959-237X} \and
Ikbeom Jang\inst{2}\corremail\orcidID{0000-0002-6901-983X}}

\authorrunning{J. Moon et al.}

\institute{Hanyang University, Seoul, Republic of Korea \and
Hankuk University of Foreign Studies, Yongin, Republic of Korea\\
\email{\{jhmoon6807, haejun\}@hanyang.ac.kr, ijang@hufs.ac.kr}}

\maketitle              

\begingroup
\renewcommand{\thefootnote}{\Letter}
\footnotetext{Corresponding authors}
\endgroup

\begin{abstract}
Accurate segmentation of thin, tortuous anatomical structures, such as retinal vessels, cerebral vasculature, and facial wrinkles, remains challenging due to low contrast, frequent discontinuities, and severe class imbalance. Although recent convolutional and Transformer-based models have improved performance, they often yield fragmented predictions and fail to recover fine branches. We propose CSWinUNETR, a general-purpose backbone for 2D and 3D thin-structure segmentation. It employs cross-shaped stripe self-attention to model long-range principal-axis context and incorporates cyclic shifts to enhance information exchange across stripes. To better preserve fine-grained details, we further introduce a detail-enhanced multi-scale self-attention module that aggregates contextual features from multi-resolution representations. In addition, we propose sparse-control dynamic snake convolution, which reconstructs reliable dense curvilinear kernels from sparsely predicted control points to better follow tortuous geometry. Extensive experiments on four benchmarks across ophthalmology, neurovascular imaging, and dermatology demonstrate that CSWinUNETR consistently outperforms state-of-the-art methods without task-specific post-processing or topology-aware losses. 
The code is available at \url{https://github.com/labhai/CSWinUNETR}.


\end{abstract}

\section{Introduction}
Segmenting thin, tortuous structures, such as retinal vessels, cerebrovascular branches, and facial wrinkles, remains a long-standing challenge across heterogeneous imaging modalities \cite{CURVSURV16,jin2022fives,moon2024facial,yang2025benchmarking}. These targets typically occupy only a minute fraction of the image domain, exhibit low contrast and frequent interruptions, and follow highly anisotropic, curvilinear paths, all of which hinder reliable segmentation \cite{FRANGI98,BOUNDARY-LOSS21,zhou2025glcp}. Despite these challenges, precise delineation of such slender structures is crucial for both clinical interpretation and the development of reliable methodological benchmarks \cite{MED-SEG-REVIEW25,yang2025benchmarking}. However, their extremely small caliber, complex branching geometry, and topological variability often confound standard architectures, leading to fragmented or missing predictions. Even state-of-the-art methods, when trained under standard protocols, often fail to preserve fine-scale details and global connectivity \cite{TOPO-LOSS20,CBDICE24,CLDICE21}.

Despite substantial progress in medical image segmentation, comparatively limited attention has been devoted to developing robust, broadly applicable architectures for delineating thin, curvilinear structures across diverse datasets and applications \cite{CURVSURV16,MED-SEG-REVIEW25}. Existing approaches targeting such structures typically rely on task-specific architectural modifications or bespoke loss functions, often combined with external priors or heavy post-processing, which hinders transfer across applications and necessitates careful parameter tuning \cite{MINPATH10,mou2021cs2,CLDICE21}. Collectively, these limitations motivate the development of a strong, general-purpose backbone that natively represents thin, tortuous structures while reducing reliance on heavy task-specific customization. 

We introduce CSWinUNETR, a segmentation backbone designed for thin, tortuous anatomical structures in 2D and 3D medical images across diverse modalities. Accurate segmentation of such structures requires two complementary capabilities: (i) orientation-aware long-range information propagation to bridge low-contrast gaps and interruptions~\cite{dong2022cswin,zhou2025glcp}, and (ii) anisotropic, trajectory-aligned integration of local evidence to preserve continuity and recover fine terminal branches~\cite{MINPATH10,qi2023dynamic}. CSWinUNETR incorporates these inductive biases in a unified architecture. To support (i), we employ Cross-Shaped Window (CSWin) self-attention~\cite{dong2022cswin} to enable efficient axis-wise global reasoning, and introduce a cyclic shift~\cite{liu2021swin} strategy to promote cross-stripe interaction and reduce boundary artifacts. To address (ii), we derive a detail-enhanced Multi-Scale Multi-Head Self-Attention (MS-MHSA) to better preserve subtle structural cues, and integrate Sparse-control Dynamic Snake Convolution (SDSConv), which aggregates reliable curvilinear features through sparse control-trajectory construction. 

Our contributions include:
(1) We propose CSWinUNETR, a general-purpose 2D/3D backbone tailored for segmenting thin, tortuous anatomical structures.
(2) We introduce shifted CSWin self-attention with multi-scale contextual modeling to enhance long-range dependency reasoning while preserving fine-grained details.
(3) We incorporate SDSConv to enable stable, trajectory-aligned local feature aggregation, strengthening geometry-aware feature representations.
(4) Extensive experiments on four heterogeneous benchmarks demonstrate consistent improvements over representative baselines, supported by both quantitative and qualitative evaluations.

\section{Method}
\label{sec:method}
\begin{figure*}[t]
  \centering
  \includegraphics[width=1\linewidth]{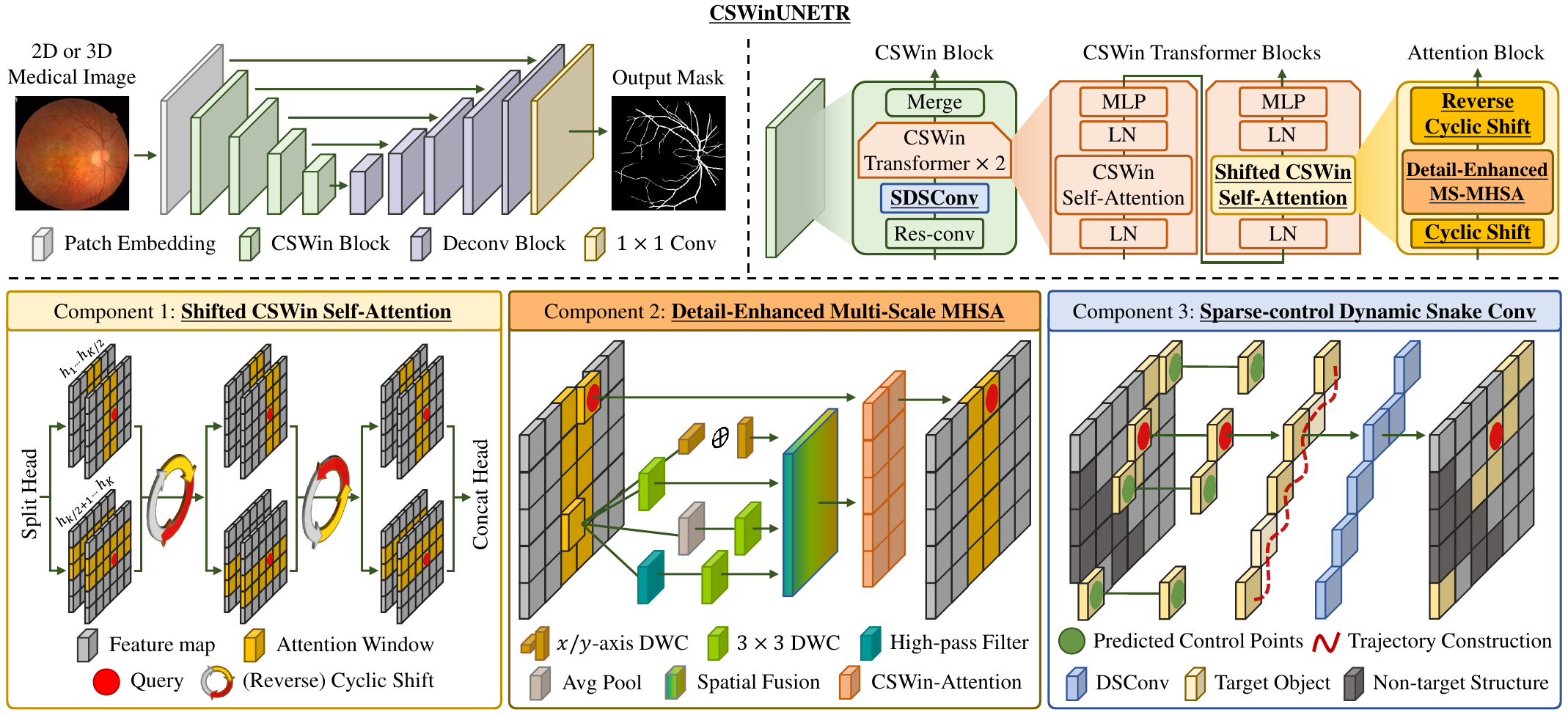}
  \caption{Overview of the proposed CSWinUNETR for segmenting thin, tortuous anatomical structures in 2D and 3D medical images. The architecture comprises three core components. (1) Shifted CSWin Self-Attention captures orientation-aware long-range dependencies via cross-shaped stripe attention with cyclic shifts. (2) Detail-Enhanced MS-MHSA fuses contextual information across multi-scale features to better preserve fine-grained structural details. (3) SDSConv encourages continuity along tortuous trajectories by constructing reliable, geometry-aligned curvilinear responses from sparsely predicted control points. Newly introduced components are \textbf{\underline{highlighted}}.}
  \label{fig:arch}
\end{figure*}

An overview of the proposed method is shown in \cref{fig:arch}. We first describe the overall architecture of CSWinUNETR (\cref{sec:backbone}), and then present the proposed core module, SDSConv (\cref{sec:sdsconv}). Although CSWinUNETR is applicable to both 2D and 3D inputs, we describe the method in the 2D setting for clarity.

\subsection{CSWinUNETR: A Backbone for Thin-Structure Segmentation}
\label{sec:backbone}
A key observation in thin-structure segmentation is that long-range dependencies in curvilinear anatomy tend to propagate predominantly along principal directions~\cite{CURVSURV16,FRANGI98}. Accordingly, axis-aligned, elongated context windows provide an effective inductive bias for aggregating non-local cues.  

Motivated by this, we adopt CSWin self-attention~\cite{dong2022cswin} to model long-range context efficiently. Given an input medical image $\mathcal{I}\in\mathbb{R}^{C_{\text{in}}\times H\times W}$, we first extract tokens using a convolutional embedding~\cite{wu2021cvt} with a $7{\times}7$ kernel and stride $2$, producing $X^{(0)}\in\mathbb{R}^{C_0\times \frac{H}{2}\times \frac{W}{2}}$. The encoder comprises four stages of CSWin blocks. For stage index $t\in\{1,2,3,4\}$, the block takes $X^{(t-1)}$ as input, applies a residual convolutional refinement, followed by two consecutive CSWin Transformer blocks, and then downsamples features through a merge layer implemented by a $3{\times}3$ convolution with stride $2$. This merge operation halves the spatial resolution and doubles the channel dimension, yielding $X^{(t)}\in\mathbb{R}^{C_t\times \frac{H}{2^{t+1}}\times \frac{W}{2^{t+1}}}$ where $C_t=2^t C_0$. For the decoder, we employ a SwinUNETR-style deconvolutional pathway~\cite{he2023swinunetr} to progressively upsample and fuse multi-scale encoder features, producing the predicted segmentation mask.

Within each CSWin block, both CSWin Transformer blocks use CSWin self-attention. Specifically, the attention heads are partitioned into two groups corresponding to vertical and horizontal stripes, and self-attention is computed independently within the associated axis-aligned stripe windows. We further apply a cyclic shift~\cite{liu2021swin} in the second CSWin Transformer block to alleviate stripe-wise isolation and boundary artifacts. With this shift, tokens near stripe boundaries are reassigned to different stripe neighborhoods, facilitating cross-stripe interactions and improving information exchange between adjacent stripes.

While stripe attention is well-suited for orientation-aware long-range aggregation, fine-grained cues of thin structures can be attenuated by stage-wise downsampling. To better preserve high-frequency details while integrating multi-scale context, we propose a detail-enhanced MS-MHSA module, inspired by multi-branch attention designs~\cite{fu2022incepformer}. Given an input feature map $X^{(t)}$, we construct several lightweight depthwise convolution (DWC) branches that operate in parallel. The axial strip branch applies a separable strip DWC along the horizontal and vertical axes and sums the two responses to reduce over-smoothing. The local-refinement branch captures regional patterns using a $3\times 3$ DWC, and the semantic-context branch models coarser context by applying average pooling (AvgPool) followed by a $3\times 3$ DWC. In addition, we incorporate a high-pass detail branch to explicitly enhance fine structures by computing $X^{(t)}-\mathrm{AvgPool}_{3\times 3}(X^{(t)})$ and refining the resulting features with a $3\times 3$ DWC.

Following multi-scale enhancement, we employ a spatial fusion gate to adaptively combine the branch outputs. Specifically, a $1\times 1$ convolution predicts per-location branch logits, which are normalized by a softmax across branches to obtain spatially varying fusion weights. The fused feature map $\widetilde{X}^{(t)}$ is then computed as the weighted sum of the branch responses. Next, we construct the attention triplet $(Q_a, K_a, V_a)$ using pointwise linear projections at each spatial location. 
The $Q_a$ is projected from the original feature map $X^{(t)}$, whereas the $K_a$ and $V_a$ are projected from the fused feature map $\widetilde{X}^{(t)}$. Given $(Q_a,K_a,V_a)$, stripe attention for each orientation branch $a\in\{\mathrm{v},\mathrm{h}\}$ is computed as:
\begin{equation}
\label{eq:cswin-attn}
O_{a}=\mathrm{Softmax}\left(
\frac{Q_{a} K_{a}^\top}{\sqrt{d}} + M_{a}
\right)V_{a} + \mathrm{LePE}(V_a),
\end{equation}
where $d$ denotes the per-head feature dimension, $M_a$ represents the (shifted) stripe attention mask, and $\mathrm{LePE}(\cdot)$ is a local positional encoding implemented by a $3{\times}3$ DWC.

\subsection{Sparse-control Dynamic Snake Convolution}
\label{sec:sdsconv}
Segmenting thin, tortuous anatomy requires anisotropic evidence aggregation that can follow curved trajectories~\cite{MINPATH10,FRANGI98}. DSConv~\cite{qi2023dynamic} partially addresses this need by deforming an axial convolutional kernel using location-adaptive offsets. However, DSConv relies on cumulative offset accumulation along the sampling path, which can amplify local prediction errors and propagate them through the entire trajectory. This issue is exacerbated when a small segment of the target structure is corrupted by non-target factors (e.g., low-contrast gaps, imaging artifacts, or occlusions), in which case subsequent sampling locations may progressively drift away from the true anatomy. 

To obtain a more stable curvilinear operator, we propose an SDSConv, which predicts sparse control points and constructs a dense snake-like kernel via trajectory construction, avoiding cumulative drift. Given the feature map $F=X^{(t)}$ at encoder level $t$, SDSConv constructs a curvilinear sampling kernel of length $P$ over a uniform axial support $p_k=k-c$, where $c=\lfloor P/2\rfloor$ and $\xi_k=p_k/c\in[-1,1]$. 
Instead of regressing $P$ independent offsets at each spatial location, SDSConv predicts $M<P$ sparse control points, comprising lateral offsets $\{\Delta_m^x,\Delta_m^y\}_{m=1}^{M}$ and learnable control-point locations $\{\mu_m\}_{m=1}^{M}\subset[-1,1]$ along the axial coordinate. 
To obtain a dense curvilinear trajectory from these sparse control points, we interpolate offsets along the kernel using Gaussian radial basis functions (RBF): 
\begin{equation}
o_u(k)=\sum_{m=1}^{M}\alpha_{m,k}\,\Delta_m^{u},\quad 
\alpha_{m,k}=\mathrm{Softmax}_m\!\left(-\frac{(\xi_k-\mu_m)^2}{2\sigma^2}\right),\quad u\in\{x,y\},
\end{equation}
where $\sigma$ denotes the RBF bandwidth. We further apply center anchoring to prevent global drift.
To stabilize deformation near the anchor while permitting larger curvature toward the kernel ends, we introduce a learnable distance-ramped scope: 
\begin{equation}
\hat{o}_u(k)=s_a(|\xi_k|)\,o_u(k),\qquad a\in\{\mathrm{v},\mathrm{h}\},
\end{equation}
where $s_a(\cdot)$ is a bounded, monotonic ramp learned for each axis branch. 
This sparse, non-cumulative trajectory construction yields a dense curvilinear kernel in a fully parallel manner, mitigating error accumulation and improving robustness under unreliable local evidence.

For each spatial location $(h,w)$, SDSConv instantiates two serpentine sampling trajectories by traversing the axial support and applying the corresponding lateral displacements $\hat{o}_x$ or $\hat{o}_y$. 
The feature map $F$ is sampled along these trajectories via bilinear interpolation, and the resulting $P$ samples are aggregated using axis-aligned DWC. 
The two axis-specific responses are then fused and injected into the CSWin blocks as a curvilinear residual prior.

\begin{table}[t]
\centering
\caption{Quantitative comparison on 2D thin-structure segmentation.}
\label{tab:benchmark_2D}

\fontsize{8pt}{9.95pt}\selectfont

\begin{tabular}{@{}l|cccc|cccc|c@{}}
\hline
\multirow{2}{*}{\textBF{Method}}
& \multicolumn{4}{c|}{\textBF{FIVES (2D)}}
& \multicolumn{4}{c|}{\textBF{FFHQ-Wrinkle (2D)}}
& \multirow{2}{*}{\textBF{\#P}} \\
\cline{2-5} \cline{6-9}
& \textBF{Dice}$\uparrow$ & \textBF{clDice}$\uparrow$ & $\beta$$\downarrow$ & \textBF{HD95}$\downarrow$
& \textBF{Dice}$\uparrow$ & \textBF{clDice}$\uparrow$ & $\beta$$\downarrow$ & \textBF{HD95}$\downarrow$
& \\
\hline

nnUNetv2~\cite{isensee2021nnu} 
& \sig{89.26}{*} & \sig{87.96}{*} & \sig{37.62}{*} & \sig{33.75}{*}
& \sig{63.27}{*} & \sig{70.69}{*} & \sig{\textBF{11.60}}{} & \sig{71.89}{*}
& \multirow{5}{*}{46M} \\

+clDice~\cite{CLDICE21} 
& \sig{89.88}{*} & \sig{90.14}{*} & \sig{36.63}{*} & \sig{29.98}{*}
& \sig{64.14}{*} & \sig{70.87}{*} & \sig{12.14}{*} & \sig{66.26}{*}
& \\

+cbDice~\cite{CBDICE24} 
& \sig{89.40}{*} & \sig{88.83}{*} & \sig{36.78}{*} & \sig{35.69}{*}
& \sig{61.96}{*} & \sig{67.47}{*} & \sig{15.53}{*} & \sig{73.79}{*}
& \\

+ske-recall~\cite{kirchhoff2024skeleton} 
& \sig{89.42}{*} & \sig{89.66}{*} & \sig{35.62}{*} & \sig{30.75}{*}
& \sig{\underline{64.40}}{} & \sig{\underline{71.90}}{*} & \sig{12.09}{*} & \sig{65.78}{*}
& \\

+GLCP~\cite{zhou2025glcp}
& \sig{88.56}{*} & \sig{89.01}{*} & \sig{48.95}{*} & \sig{39.09}{*}
& \sig{62.91}{*} & \sig{70.25}{*} & \sig{12.03}{*} & \sig{65.88}{*}
&  \\

\hline

UNETR~\cite{hatamizadeh2022unetr}
& \sig{87.65}{*} & \sig{88.50}{*} & \sig{56.02}{*} & \sig{36.18}{*}
& \sig{55.62}{*} & \sig{62.81}{*} & \sig{17.79}{*} & \sig{99.96}{*}
& 88M \\

SwinUNETR~\cite{hatamizadeh2021swin}
& \sig{89.35}{*} & \sig{90.25}{*} & \sig{41.46}{*} & \sig{\underline{26.96}}{}
& \sig{62.64}{*} & \sig{70.28}{*} & \sig{12.25}{*} & \sig{65.75}{*}
& 25M \\ %

SwinUNETRv2~\cite{he2023swinunetr}
& \sig{\underline{89.90}}{*} & \sig{90.20}{*} & \sig{39.37}{*} & \sig{27.18}{*}
& \sig{63.57}{*} & \sig{71.48}{} & \sig{11.83}{} & \sig{65.89}{*}
& 29M \\ %

\hline

IncepFormer~\cite{fu2022incepformer}
& \sig{89.57}{*} & \sig{89.89}{*} & \sig{56.27}{*} & \sig{28.59}{*}
& \sig{62.46}{*} & \sig{69.69}{*} & \sig{11.70}{} & \sig{63.43}{*}
& 15M \\ 

DSCNet~\cite{qi2023dynamic}
& \sig{88.22}{*} & \sig{88.46}{*} & \sig{63.02}{*} & \sig{30.43}{*}
& \sig{61.40}{*} & \sig{69.47}{*} & \sig{11.95}{*} & \sig{66.12}{*}
& 2M \\ %

CSWin-UNET~\cite{liu2025cswin}
& \sig{89.80}{*} & \sig{\underline{90.83}}{} & \sig{\underline{34.34}}{} & \sig{27.00}{}
& \sig{62.57}{*} & \sig{70.00}{*} & \sig{11.91}{} & \sig{\underline{63.18}}{}
& 24M \\ %

\hline

CS$^{2}$-Net~\cite{mou2021cs2}
& \sig{87.69}{*} & \sig{88.75}{*} & \sig{60.66}{*} & \sig{36.21}{*}
& \sig{59.91}{*} & \sig{67.54}{*} & \sig{12.10}{*} & \sig{72.18}{*}
& 8M \\ 

SGAT-Net~\cite{lin2023stimulus}
& \sig{88.95}{*} & \sig{89.42}{*} & \sig{42.68}{*} & \sig{33.43}{*}
& \sig{55.27}{*} & \sig{55.81}{*} & \sig{18.03}{*} & \sig{94.82}{*}
& 8M \\ 

TMP+U-Net~\cite{moon2024facial}
& \sig{86.64}{*} & \sig{87.17}{*} & \sig{64.56}{*} & \sig{37.67}{*}
& \sig{61.44}{*} & \sig{68.67}{*} & \sig{12.27}{*} & \sig{85.11}{*}
& 17M \\

\hline

\rowcolor{gray!30} CSWinUNETR (Ours) 
& \textBF{90.71} & \textBF{90.88} & \textBF{33.07} & \textBF{25.62}
& \textBF{64.75} & \textBF{72.18} & \underline{11.67} & \textBF{61.76}
& 31M \\
\hline
\end{tabular}

\end{table}

\begin{table}[t]
\caption{Quantitative comparison on 3D thin-structure segmentation.}
\label{tab:benchmark_3D}
\centering

\fontsize{8pt}{10pt}\selectfont

\begin{tabular}{@{}l|cccc|cccc|c@{}}
\hline
\multirow{2}{*}{\textBF{Method}}
& \multicolumn{4}{c|}{\textBF{TopCoW-MRA (3D)}}
& \multicolumn{4}{c|}{\textBF{TopCoW-CTA (3D)}}
& \multirow{2}{*}{\textBF{\#P}} \\
\cline{2-5} \cline{6-9}
& \textBF{Dice}$\uparrow$ & \textBF{clDice}$\uparrow$ & $\beta$$\downarrow$ & \textBF{HD95}$\downarrow$
& \textBF{Dice}$\uparrow$ & \textBF{clDice}$\uparrow$ & $\beta$$\downarrow$ & \textBF{HD95}$\downarrow$
& \\
\hline

nnUNetv2~\cite{isensee2021nnu}
& \sig{80.84}{*} & \sig{89.45}{*} & \sig{0.74}{*} & \sig{7.78}{*}
& \sig{75.23}{*} & \sig{83.89}{*} & \sig{0.56}{*} & \sig{8.14}{*}
& \multirow{5}{*}{31M} \\

+clDice~\cite{CLDICE21}
& \sig{\underline{82.65}}{*} & \sig{\underline{91.64}}{*} & \sig{0.68}{*} & \sig{6.64}{*}
& \sig{76.14}{*} & \sig{84.75}{*} & \sig{0.48}{*} & \sig{7.89}{*}
& \\

+cbDice~\cite{CBDICE24}
& \sig{82.49}{*} & \sig{91.32}{*} & \sig{\underline{0.66}}{} & \sig{6.85}{*}
& \sig{\underline{76.43}}{*} & \sig{86.35}{*} & \sig{0.45}{*} & \sig{7.35}{}
& \\

+ske-recall~\cite{kirchhoff2024skeleton}
& \sig{81.94}{*} & \sig{91.28}{*} & \sig{0.71}{*} & \sig{7.46}{*}
& \sig{75.86}{*} & \sig{82.64}{*} & \sig{0.54}{*} & \sig{8.02}{*}
& \\

+GLCP~\cite{zhou2025glcp}
& \sig{81.78}{*} & \sig{90.68}{*} & \sig{0.77}{*} & \sig{6.89}{*}
& \sig{74.69}{*} & \sig{\underline{86.68}}{} & \sig{0.59}{*} & \sig{7.64}{*}
& \\

\hline

UNETR~\cite{hatamizadeh2022unetr}
& \sig{78.80}{*} & \sig{84.56}{*} & \sig{0.94}{*} & \sig{7.55}{*}
& \sig{65.84}{*} & \sig{73.13}{*} & \sig{0.92}{*} & \sig{9.48}{*}
& 93M \\

SwinUNETR~\cite{hatamizadeh2021swin}
& \sig{80.98}{*} & \sig{84.54}{*} & \sig{0.71}{*} & \sig{7.69}{*}
& \sig{72.48}{*} & \sig{82.40}{*} & \sig{0.62}{*} & \sig{\underline{6.89}}{}
& 62M \\

SwinUNETRv2~\cite{he2023swinunetr}
& \sig{79.96}{*} & \sig{85.38}{*} & \sig{0.74}{*} & \sig{7.54}{*}
& \sig{72.87}{*} & \sig{84.29}{*} & \sig{\underline{0.44}}{} & \sig{8.18}{*}
& 73M \\

\hline

DSCNet~\cite{qi2023dynamic}
& \sig{82.49}{*} & \sig{90.55}{*} & \sig{0.67}{} & \sig{\underline{6.16}}{}
& \sig{73.32}{*} & \sig{83.53}{*} & \sig{0.69}{*} & \sig{8.45}{*}
& 8M \\

\hline

CS$^{2}$-Net~\cite{mou2021cs2}
& \sig{80.30}{*} & \sig{84.54}{*} & \sig{0.98}{*} & \sig{6.51}{*}
& \sig{73.38}{*} & \sig{84.32}{*} & \sig{0.53}{*} & \sig{7.77}{*}
& 23M \\

ER-Net~\cite{xu2023ernet}
& \sig{73.86}{*} & \sig{78.13}{*} & \sig{0.87}{*} & \sig{8.76}{*}
& \sig{73.50}{*} & \sig{83.46}{*} & \sig{0.91}{*} & \sig{8.73}{*}
& 5M \\

\hline

\rowcolor{gray!30} CSWinUNETR (Ours)
& \textBF{84.74} & \textBF{93.55} & \textBF{0.58} & \textBF{4.72}
& \textBF{77.06} & \textBF{90.71} & \textBF{0.36} & \textBF{6.11}
& 81M \\
\hline
\end{tabular}

\end{table}

\subsection{Optimization and Implementation Details}
To assess generality, we adopt an identical network architecture across all datasets, modifying only the settings that depend on the input dimensionality when switching between 2D and 3D.
After convolutional tokenization, the base embedding dimension is fixed to $C_0=48$.
The number of attention heads is set to $(2,4,8,16)$ for 2D inputs and $(3,6,12,24)$ for 3D inputs.
For the SDSConv, the kernel length $P$ is set to $(11,9,7,5)$ from shallow to deep stages, and the number of predicted sparse control points is defined as $M=\lfloor P/2\rfloor$.
Model training is performed using a composite objective combining Dice and cross-entropy losses.

\section{Experiments}
\label{sec:experiment}

\subsection{Datasets and Evaluation metrics}
We evaluate four benchmarks for thin, tortuous-structure segmentation across ophthalmology, neurovascular imaging, and dermatology. 
FIVES~\cite{jin2022fives} comprises 800 fundus images at \(2048\times2048\) resolution with binary vessel annotations, split into 500/100/200 images for training/validation/testing. 
TopCoW~\cite{yang2025benchmarking} includes paired 3D CTA and MRA Circle-of-Willis volumes with multiclass annotations for 13 vessel classes.
We use the public 125-patient cohort and adopt a 100/12/13 patient-level split for training/validation/testing. 
FFHQ-Wrinkle~\cite{moon2024facial} provides binary wrinkle masks for 1{,}000 facial images at \(1024\times1024\) resolution, with an 800/100/100 training/validation/testing split. 
Segmentation performance is measured using overlap metrics (Dice~\cite{dice1945measures} and clDice~\cite{CLDICE21}), topology error (Betti Error~\cite{hu2019topology}, \(\beta=\beta_0+\beta_1\)), and boundary distance error (95th-percentile Hausdorff Distance; HD95~\cite{taha2015metrics}).

\subsection{Comparison Results and Ablation Study}
\label{sec:results-quant}
To evaluate CSWinUNETR on thin-structure segmentation, we first consider general-purpose medical segmentation backbones, including nnUNetv2~\cite{isensee2021nnu}, UNETR~\cite{hatamizadeh2022unetr}, SwinUNETR~\cite{hatamizadeh2021swin}, and SwinUNETRv2~\cite{he2023swinunetr}. We then include methods that incorporate components closely related to our design, namely IncepFormer~\cite{fu2022incepformer}, DSCNet~\cite{qi2023dynamic}, and CSWin-UNET~\cite{liu2025cswin}. We also compare with thin-structure-oriented methods, including CS$^{2}$-Net~\cite{mou2021cs2}, SGAT-Net~\cite{lin2023stimulus}, Texture-Map Pretraining (TMP)+U-Net~\cite{moon2024facial}, GLCP~\cite{zhou2025glcp}, and ER-Net~\cite{xu2023ernet}. In addition, we retrain nnUNetv2 using clDice~\cite{CLDICE21}, cbDice~\cite{CBDICE24}, and skeleton-recall (ske-recall) loss~\cite{kirchhoff2024skeleton}. For statistical analysis, we use a paired $t$-test with the Benjamini–Hochberg correction. In all tables, statistical significance is indicated by $^{*}$ ($p<0.05$). \#P denotes the number of trainable parameters; for nnU-Net-based methods, \#P varies with the configuration and is therefore reported as a range. The best and second-best results are highlighted in \textbf{bold} and \underline{underlined}.

\begin{figure*}[t]
  \centering
  \includegraphics[width=\linewidth]{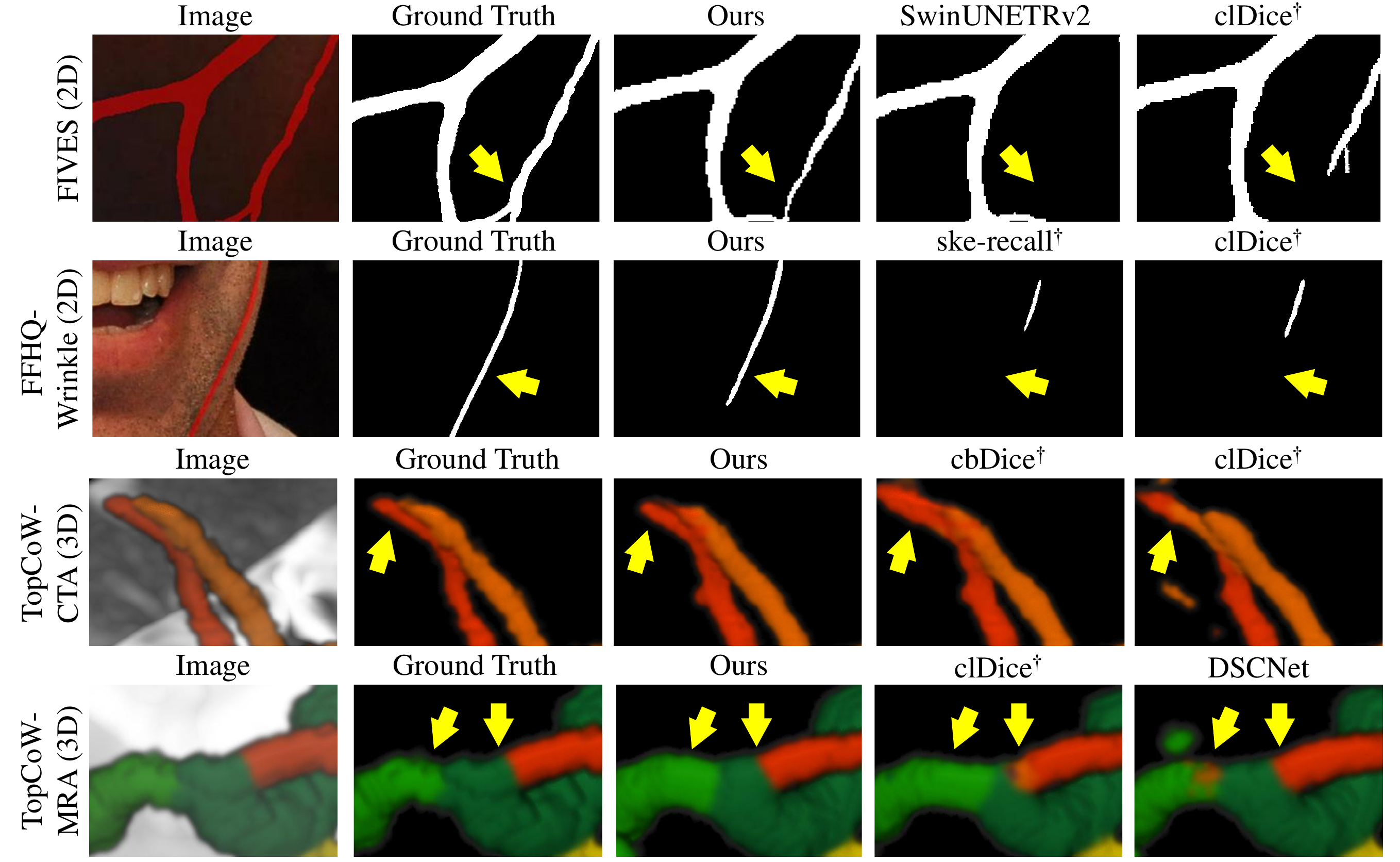}
  \caption{Qualitative comparison of 2D and 3D thin-structure segmentation. From left to right: the input image overlaid with the ground truth, the ground truth, our CSWinUNETR prediction, and the second- and third-ranked methods by Dice score (method names are shown above). $\dagger$ denotes methods trained with an nnUNetv2 framework. Yellow arrows highlight regions with the largest discrepancies.}
  \label{fig:qual}
\end{figure*}

\begin{table}[t]
\caption{Ablation study of the proposed core components. ``Cyc'' denotes the cyclic shift, ``MS'' is the detail-enhanced MS-MHSA, and ``SDSC'' represents the SDSConv.}
\label{tab:ablation_topcow}
\centering
\fontsize{8pt}{10pt}\selectfont
\begin{tabular}{@{}ccc|cccc|c|cccc|c@{}}
\hline
\multicolumn{3}{c|}{\textBF{Method}} &
\multicolumn{5}{c|}{\textBF{TopCoW-MRA (3D)}} &
\multicolumn{5}{c}{\textBF{FFHQ-Wrinkle (2D)}} \\
\cline{1-3} \cline{4-8} \cline{9-13}
\textBF{Cyc} & \textBF{MS} & \textBF{SDSC} &
\textBF{Dice}$\uparrow$ & \textBF{clDice}$\uparrow$ & $\beta$$\downarrow$ & \textBF{HD95}$\downarrow$ & \textBF{\#P} &
\textBF{Dice}$\uparrow$ & \textBF{clDice}$\uparrow$ & $\beta$$\downarrow$ & \textBF{HD95}$\downarrow$ & \textBF{\#P} \\
\hline
 &  &  & 82.81 & 91.93 & 0.77 & 6.76 & 79.9M & 64.09 & 71.55 & 11.86 & 63.89 & 30.5M \\
\checkmark &  &  & 83.39 & 92.28 & 0.74 & 5.91 & 79.9M & 64.24 & 71.78 & 11.84 & 63.61 & 30.5M \\
\checkmark & \checkmark &  & 83.96 & 92.77 & 0.64 & 5.16 & 80.0M & 64.35 & 71.94 & 11.79 & 63.29 & 30.6M \\
\rowcolor{gray!30}
\checkmark & \checkmark & \checkmark & \textBF{84.74} & \textBF{93.55} & \textBF{0.58} & \textBF{4.72} & 80.8M &
\textBF{64.75} & \textBF{72.18} & \textBF{11.67} & \textBF{61.76} & 31.1M \\
\hline
\end{tabular}
\end{table}

Tables~\ref{tab:benchmark_2D} and~\ref{tab:benchmark_3D} indicate that CSWinUNETR consistently outperforms the competing methods across all four benchmarks. In particular, the observed improvements are statistically significant ($p<0.05$) for most metrics relative to the other approaches. The qualitative comparisons in \cref{fig:qual} further corroborate these findings, showing that CSWinUNETR more faithfully reconstructs fine branches and terminal structures, while reducing false positives in non-target regions. Table~\ref{tab:ablation_topcow} indicates that adding the cyclic shift, MS-MHSA, and SDSConv leads to consistent, stepwise improvements across all evaluation metrics. In particular, the cyclic shift enhances performance without increasing the number of parameters. The addition of MS-MHSA and SDSConv further strengthens the results, with only a modest parameter overhead.

\section{Discussion and Conclusion}
The proposed CSWinUNETR consistently outperforms strong baselines across four heterogeneous benchmarks that emphasize fine-grained, tortuous anatomical structures. Notably, these benchmarks span visually and modality-diverse targets, ranging from vascular segmentation to facial wrinkle delineation. Despite this diversity, we evaluate a single end-to-end model with a largely fixed architecture and training protocol across all datasets, without task-specific post-processing or topology-aware auxiliary objectives. The consistent improvements observed in this unified setting indicate that the performance gains are more plausibly attributable to CSWinUNETR's architectural inductive biases than to extensive dataset-specific hyperparameter tuning. Taken together, these results support CSWinUNETR as a practical, off-the-shelf backbone for thin-structure segmentation across varying imaging modalities.

The qualitative results further indicate that CSWinUNETR more effectively preserves structural continuity while reducing spurious predictions outside the target anatomy. We attribute these improvements to the combined effects of orientation-aware long-range contextual modeling and trajectory-aligned feature aggregation. In particular, SDSConv constructs curvilinear sampling trajectories from a sparse set of control points and aggregates features along anatomically plausible paths. By emphasizing consistent evidence across the trajectory, this mechanism can recover continuity even when the structure appears locally disrupted due to limited spatial resolution, motion, or acquisition artifacts. This property is clinically relevant because thin, tortuous anatomies can appear fragmented in imaging despite being anatomically continuous. By producing more consistently connected segmentations, CSWinUNETR may better support downstream centerline-based analyses, including the estimation of length, branching patterns, and tortuosity.

To conclude, CSWinUNETR is a general-purpose segmentation architecture for thin, tortuous anatomical structures across heterogeneous medical imaging modalities, encompassing both 2D images and 3D volumes. Extensive experiments demonstrate that CSWinUNETR consistently outperforms existing methods, including approaches tailored to thin-structure segmentation. These results suggest that CSWinUNETR offers a principled framework for geometry-aware attention and robust curvilinear feature aggregation, enabling reliable dense prediction in challenging curvilinear segmentation scenarios.

\bibliographystyle{splncs04}
\bibliography{reference}

\end{document}